\documentclass{article}
\usepackage{spconf,amsmath,amssymb,graphicx,booktabs,array,url}
\usepackage[T1]{fontenc}
\usepackage{times}
\usepackage{xcolor}
\usepackage{flushend}
\title{\fontsize{14}{16}\selectfont Dual-Region Slice Aggregation\\for CT-Based Renal Tumor Classification}
\name{Yuan Liang\textsuperscript{1,3}, Sourav Bhattacharjee\textsuperscript{2}, and Abraham Campbell\textsuperscript{3}}
\address{{\fontsize{10}{12}\selectfont
\begin{tabular}{@{}c@{}}
\textsuperscript{1} Research Ireland Centre for Research Training in Machine Learning\\
\textsuperscript{2} School of Veterinary Medicine, University College Dublin, Dublin, Ireland\\
\textsuperscript{3} School of Computer Science, University College Dublin, Dublin, Ireland\\
yuan.liang@ucdconnect.ie
\end{tabular}}}
\begin{document}
\ninept
\maketitle

\begin{abstract}
Clear cell renal cell carcinoma (ccRCC) is the most common kidney cancer in adults. We present a region-aware architecture for CT-based ccRCC classification. Tumor and 0--10\,mm peritumoral context are pooled separately from shared frozen DINOv3 feature maps. A shared position-aware slice aggregator precedes fixed training-only principal component analysis (PCA). The resulting representations are fused with radiomics using two scalar feature gates. Evaluation on KiTS23 and 111 external cases yields internal test and external AUCs of $0.8078\pm0.0154$ and $0.9094\pm0.0015$. Compared with matched single-tumor readout, mean AUC increases by 1.50 and 2.34 percentage points, respectively. Early-PCA and reconstruction controls yield lower internal mean AUCs, while external AUC is similar for early and late PCA. These findings highlight the joint role of regional readout and slice aggregation in adapting frozen visual features to renal CT.
\end{abstract}

\section{Introduction}
\label{sec:intro}
CT-based characterization of renal masses can use information from both the tumor and its surrounding tissue~\cite{peritumor}. Pretrained visual encoders can be adapted to small labeled CT cohorts by training downstream aggregation and classification modules. DINOv3 provides dense visual features for this adaptation~\cite{dinov3}. Whole-field pooling summarizes the imaged field, whereas regional readout preserves the distinction between tumor and surrounding tissue. Slice-level features must then be combined into a patient-level representation.

Radiomics describes image intensity, texture, and shape~\cite{pyradiomics}. Previous renal studies have extracted tumor and peritumoral radiomic features for ccRCC characterization~\cite{peritumor}. We extend this spatial separation to deep representations by pooling tumor and peritumoral features from the same frozen DINOv3 maps. A shared, position-aware aggregator then forms a patient-level descriptor for each region.

The order of aggregation and dimensionality reduction determines which features are available to the trainable module. Slice-wise PCA confines aggregation to a fixed low-dimensional subspace. Aggregation before PCA permits learned transformations of the original slice features before the final compact representation is formed. This distinction remains even when the two designs share the same training-derived PCA basis.

\begin{figure*}[t]
\centering
\includegraphics[width=\textwidth]{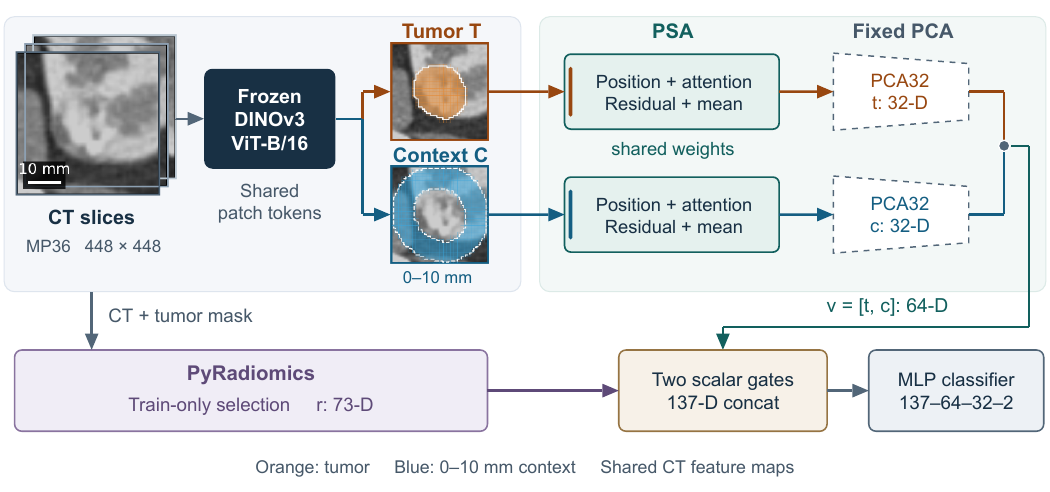}
\caption{Dual-region architecture illustrated with a KiTS23 training CT. Orange and blue overlays show tumor (T) and 0--10\,mm context (C) pooling weights on the shared patch grid. The encoder receives unmasked CT slices. T and C pass separately through the same trainable position-aware slice aggregator (PSA), followed by fixed PCA32. Two scalar gates fuse the concatenated image vector with radiomics. CT is displayed at $[-200,300]$ HU.}
\label{fig:architecture}
\end{figure*}

The proposed architecture (Fig.~\ref{fig:architecture}) combines separate tumor and peritumoral readouts with position-aware slice aggregation before fixed compression. Radiomics complements the aggregated deep representations through gated fusion. We evaluate regional readout against matched whole-field and single-tumor configurations, and examine compression order using early-PCA and input-reconstruction controls. Comparisons with radiomics, frozen 3D features, and RadioDINO assess the architecture alongside alternative representations.

\section{Method}
\label{sec:method}
\subsection{Physical sampling and regional readout}
Each CT and tumor mask is placed on a 1\,mm right--anterior--superior grid. A cube centered on the tumor bounding box uses an edge equal to the largest tumor extent plus a 10\,mm margin on each side. Twelve positions span the tumor extent in each of the axial, coronal, and sagittal planes, giving 36 slices (MP36). Each slice is sampled at $448\times448$ pixels. Intensities are clipped to $[-200,300]$ HU, mapped to $[0,1]$, repeated over three channels, and normalized with the backbone input statistics.

The frozen DINOv3 ViT-B/16 encoder produces a $28\times28$ grid of 768-dimensional patch tokens per slice. Let $f_{sp}\in\mathbb{R}^{768}$ denote token $p$ on slice $s$. T is the tumor mask. C is the non-tumor region within a physical distance of 10\,mm, intersected with valid CT support. This distance-based context may include tissue outside the renal parenchyma. Fractional region occupancy is mapped to the patch grid by area integration. The regional feature is
\begin{equation}
 x_s^{(r)}=\frac{\sum_p a_{sp}^{(r)}f_{sp}}{\sum_p a_{sp}^{(r)}},\qquad r\in\{T,C\},
 \label{eq:readout}
\end{equation}
where $a_{sp}^{(r)}$ includes region occupancy and valid image support. Whole-field readout uses valid CT support in place of the regional mask. All readouts share the same CT slices and backbone tokens.

\subsection{Aggregation before fixed PCA}
Training-set statistics standardize each regional slice feature to $\tilde{x}_s^{(r)}$. A learned input projection $W_{\rm in}$ maps 768 dimensions to 64. The fixed physical encoding $e$ identifies the anatomical axis, distance from the tumor center, and normalized position within its extent. The shared aggregator $\mathcal{T}_{\theta}$ is one pre-normalized Transformer layer with four attention heads, a 128-dimensional feed-forward sublayer, GELU, and dropout 0.1. Its contextualized slice output $h_s^{(r)}$ is projected back to 768 dimensions by $W_{\rm out}$ and added to the standardized input. The binary mask $m_s^{(r)}$ marks valid regional support, so averaging includes only valid slices:
\begin{align}
 h^{(r)}&=\mathcal{T}_{\theta}(W_{\rm in}\tilde{x}^{(r)}+e,m^{(r)}),\nonumber\\
 u^{(r)}&=\frac{\sum_s m_s^{(r)}(\tilde{x}_s^{(r)}+W_{\rm out}h_s^{(r)})}
 {\sum_s m_s^{(r)}},\quad z^{(r)}=\mathcal{P}(u^{(r)}).
 \label{eq:aggregate}
\end{align}
Here, $u^{(r)}$ is the pooled patient-level regional feature and $z^{(r)}$ is its 32-dimensional PCA representation. Bias terms are omitted for clarity. T and C are processed independently by a shared slice aggregator.

$\mathcal{P}$ is a fixed 32-dimensional affine PCA transform followed by component standardization. The PCA basis is fitted to unstandardized whole-field training features, then expressed in the standardized feature coordinates used by the aggregator. The transform remains fixed during downstream training. The two regional representations form the image vector $v=[z^{(T)};z^{(C)}]\in\mathbb{R}^{64}$.

\subsection{Radiomics and fusion}
We extract 386 original and Laplacian-of-Gaussian features with PyRadiomics~\cite{pyradiomics}, using Image Biomarker Standardisation Initiative (IBSI)-based feature definitions~\cite{ibsi}. Stability filtering uses ICC(3,1)$\geq0.75$ on 160 training cases with three annotations. Training-only filtering then removes features below the 20th percentile of median absolute deviation, prunes absolute Pearson correlations above 0.95, and selects representatives from average-linkage clusters at distance 0.2. This yields 73 standardized features, $r\in\mathbb{R}^{73}$, used unchanged by the compared configurations.

We fuse radiomics and deep CT features using multiplicative gating~\cite{gmu}. Two scalar gates are computed from the concatenated feature vectors using an affine layer and a sigmoid scaled by two. The gates weight the radiomics vector and the entire T/C image vector, respectively. Zero-initialized affine weights and biases set both gates to one initially. The weighted vectors are concatenated and passed to a $137\rightarrow64\rightarrow32\rightarrow2$ MLP with GELU and dropout 0.1 to predict the class logits. The backbone and preprocessing remain frozen. Only aggregation, gates, and the classifier are trained.

\section{Experimental design}
\label{sec:experiments}
\subsection{Cohorts and evaluation}
The analysis uses a 396-case KiTS23-derived cohort~\cite{kits23}, divided into training, validation, and internal test sets in approximately 60\%/25\%/15\% proportions. Approximately 70\% of cases in each subset are ccRCC. The external cohort contains 111 TCGA cases obtained through TCIA~\cite{tcia}, of which 76.6\% are ccRCC. External tumor masks are drawn from two community annotation resources: RCC-AID v2~\cite{rccaid} and BAMF-AIMI~\cite{aimi}. All methods are evaluated on the same patient cohorts and diagnostic labels.

All reported neural configurations are evaluated using three random seeds. Regional DINO models and their ablations use AdamW with learning rate $10^{-3}$, zero weight decay, batch size 16, gradient-norm clipping at 1, and class-weighted cross-entropy. For each run, we retain the model with the highest validation AUC, choosing the earliest epoch in a tie. PCA, scalers, and radiomics selection are fitted on training data only. Baseline hyperparameters follow their original validation-based selection.

AUC is reported as the mean and sample SD across three seeds. Comparisons use differences between mean AUCs.

\subsection{Baselines and ablation studies}
Mean averages valid, standardized slice features without a Transformer or PCA. PSA--PCA denotes position-aware slice aggregation followed by fixed PCA (Eq.~\eqref{eq:aggregate}). Their comparison assesses the combined aggregation and compression design. Table~\ref{tab:main} compares whole-field, single-tumor, and T/C readouts with the same feature-fusion gate. Changing the number of regions also changes the classifier input width.

Radiomics-only logistic regression (LR) uses the same 73 radiomic features, balanced class weights, and L2 regularization. Validation selects $C=0.01$ from $\{10^{-3},10^{-2},10^{-1},1,10,100\}$. Frozen MedicalNet 3D ResNet18 plus radiomics~\cite{med3d} provides a 3D baseline. The matched RadioDINO-B/16 control~\cite{radiodino} uses identical sampling, regional readout, aggregation, fusion, and training settings. Backbone-specific image scaling and PCA are fitted on the training set.

Table~\ref{tab:ablation} isolates radiomics, slice interaction, compression order, and gated fusion. Radiomics is ablated by zeroing its input while retaining the classifier. The no-interaction control replaces attention with a token-wise $64\rightarrow128\rightarrow64$ GELU MLP, keeping all other operations. Early PCA reduces slice features to 32 dimensions before aggregation. PCA reconstruction applies fixed PCA32 and its inverse before PSA--PCA, preserving the architecture and parameter count while restricting the input subspace. Direct concatenation joins radiomics and deep CT features without gate weighting.

\begin{table}[t]
\centering
\caption{Mean AUC $\pm$ sample SD over three random seeds. Mean: slice mean pooling. PSA--PCA: position-aware slice aggregation followed by fixed PCA. R: radiomics. Regional/whole models use DINOv3 unless marked RadioDINO.}
\label{tab:main}
\setlength{\tabcolsep}{2.5pt}
\renewcommand{\arraystretch}{1.12}
\begin{tabular}{lcc}
\toprule
Method & Test ($n=60$) & External ($n=111$) \\
\midrule
Radiomics LR$^{\dagger}$ & 0.7513 $\pm$ 0.0000 & 0.8597 $\pm$ 0.0000 \\
\specialrule{0.25pt}{2pt}{2pt}
3D ResNet18 + R & 0.6887 $\pm$ 0.0188 & 0.7814 $\pm$ 0.0270 \\
\specialrule{0.25pt}{2pt}{2pt}
\begin{tabular}[c]{@{}l@{}}RadioDINO-B/16\\T/C, PSA--PCA\\+ R + gate\end{tabular} & 0.7368 $\pm$ 0.0426 & 0.8302 $\pm$ 0.0294 \\
\specialrule{0.25pt}{2pt}{2pt}
\begin{tabular}[c]{@{}l@{}}Whole, PSA--PCA\\+ R + gate\end{tabular} & 0.7363 $\pm$ 0.0163 & 0.8603 $\pm$ 0.0248 \\
\specialrule{0.25pt}{2pt}{2pt}
\begin{tabular}[c]{@{}l@{}}T, Mean\\+ R + gate\end{tabular} & 0.7923 $\pm$ 0.0145 & 0.8929 $\pm$ 0.0079 \\
\specialrule{0.25pt}{2pt}{2pt}
\begin{tabular}[c]{@{}l@{}}T, PSA--PCA\\+ R + gate\end{tabular} & 0.7928 $\pm$ 0.0205 & 0.8860 $\pm$ 0.0122 \\
\specialrule{0.25pt}{2pt}{2pt}
\begin{tabular}[c]{@{}l@{}}T/C, Mean\\+ R + gate\end{tabular} & 0.7835 $\pm$ 0.0159 & 0.8971 $\pm$ 0.0088 \\
\midrule
\begin{tabular}[c]{@{}l@{}}\textbf{T/C, PSA--PCA}\\\textbf{+ R + gate}\end{tabular} & \textbf{0.8078 \boldmath$\pm$ 0.0154} & \textbf{0.9094 \boldmath$\pm$ 0.0015} \\
\bottomrule
\end{tabular}
\par\smallskip\parbox{\columnwidth}{\normalsize $^{\dagger}$Deterministic LR refits yield identical predictions and zero SD.}
\end{table}

\section{Results}
\label{sec:results}
\begin{table*}[t]
\centering
\caption{Ablation results as mean AUC $\pm$ sample SD over three random seeds. $\Delta$ is the full-model mean AUC minus the control mean AUC. Positive values favor the full model.}
\label{tab:ablation}
\setlength{\tabcolsep}{7pt}
\renewcommand{\arraystretch}{1.12}
\begin{tabular}{lcccc}
\toprule
Configuration & Test AUC & $\Delta$ Test & External AUC & $\Delta$ External \\
\midrule
Full model & 0.8078 $\pm$ 0.0154 & --- & 0.9094 $\pm$ 0.0015 & --- \\
Without radiomics & 0.7892 $\pm$ 0.0088 & +0.0185 & 0.8599 $\pm$ 0.0124 & +0.0495 \\
Without slice interaction & 0.7879 $\pm$ 0.0135 & +0.0198 & 0.8747 $\pm$ 0.0223 & +0.0347 \\
PCA before aggregation & 0.7474 $\pm$ 0.0103 & +0.0604 & 0.9103 $\pm$ 0.0117 & -0.0009 \\
\begin{tabular}[c]{@{}l@{}}PCA reconstruction\\before PSA--PCA\end{tabular} & 0.7795 $\pm$ 0.0213 & +0.0282 & 0.8992 $\pm$ 0.0190 & +0.0101 \\
Direct concatenation & 0.7937 $\pm$ 0.0119 & +0.0141 & 0.9101 $\pm$ 0.0050 & -0.0008 \\
\bottomrule
\end{tabular}

\end{table*}

\subsection{Regional readout and aggregation}
The main model achieves test AUC $0.8078\pm0.0154$ and external AUC $0.9094\pm0.0015$ (Table~\ref{tab:main}). Relative to the whole-field PSA--PCA configuration, mean AUC increases by 0.0714 internally and 0.0490 externally.

Relative to the single-tumor PSA--PCA configuration, the corresponding increases are 0.0150 and 0.0234. Relative to the single-tumor Mean configuration with gating, mean AUC increases by 0.0155 internally and 0.0164 externally.

The regional contrast differs between aggregation strategies. With Mean, T/C readout has lower mean test AUC than T alone (0.7835 versus 0.7923). Its external mean is higher by 0.0042. Within T/C, PSA--PCA exceeds Mean by 0.0243 internally and 0.0122 externally.

The matched RadioDINO control yields mean test and external AUCs of 0.7368 and 0.8302, respectively. DINOv3 has higher mean AUCs in both cohorts, with differences of 0.0710 internally and 0.0792 externally.

\subsection{PCA placement and input reconstruction}
In Table~\ref{tab:ablation}, early PCA yields mean test AUC 0.7474, lower than the full model by 0.0604. The reconstruction control yields a mean test AUC of 0.7795, with the full model higher by 0.0282.

Externally, early and late PCA yield mean AUCs of 0.9103 and 0.9094, respectively. The reconstruction control reaches 0.8992. The full-model mean is higher by 0.0101.

\subsection{Radiomics, interaction, and fusion}
Removing radiomics lowers mean AUC by 0.0185 internally and 0.0495 externally (Table~\ref{tab:ablation}). Replacing slice interaction lowers means by 0.0198 and 0.0347.

Gating increases mean internal test AUC by 0.0141 relative to direct concatenation. The external mean difference is $-0.0008$.

\section{Discussion and conclusion}
Separate tumor and peritumoral streams retain regional information through slice aggregation while sharing encoder features and aggregation weights. Under PSA--PCA, T/C readout yields higher mean AUC than matched whole-field and single-tumor readouts in both cohorts. Regional pooling separates tumor appearance from nearby anatomy, whereas PSA models relationships among slices within each region. Sharing aggregation weights preserves separate patient-level descriptors under a common transformation. In the internal cohort, adding context reduces mean AUC under mean pooling but increases it under PSA--PCA (Table~\ref{tab:main}). This pattern suggests that the contribution of context depends on how regional slice features are combined.

The lower internal mean AUC of the reconstruction control is consistent with a benefit from learning transformations of the original slice features before fixed compression. The reconstruction control preserves the aggregator and classifier architecture while restricting the input to the PCA subspace. PCA reconstruction maps the compressed features back to 768 dimensions without recovering components discarded by the fixed projection. Early PCA also reduces the aggregator input and output dimensions. The reconstruction comparison therefore evaluates input restriction at matched model capacity. Externally, early and late PCA have similar mean AUCs, while the full model exceeds the reconstruction control. The compression-order comparison thus shows different patterns across cohorts.

The radiomics ablation and fusion comparison address different aspects of feature integration (Table~\ref{tab:ablation}). Removing radiomics lowers mean AUC in both cohorts, with a larger external decrease of 0.0495, consistent with complementary information in handcrafted descriptors and frozen deep features. Gating increases internal mean AUC by 0.0141 over direct concatenation, whereas its external mean is lower by 0.0008. Radiomics therefore contributes in both cohorts, while the gain from adaptive weighting is specific to the internal comparison.

We present a dual-region architecture that preserves separate tumor and peritumoral readouts through position-aware slice aggregation before fixed PCA, then integrates radiomics through gated fusion. The evaluated configuration achieves higher mean AUC than matched whole-field and single-tumor baselines in both cohorts. Radiomics fusion improves mean AUC in both cohorts, while the internal ablations favor aggregation before fixed compression.

\section{Compliance with ethical standards}
This retrospective study used publicly available KiTS23 and TCGA/\allowbreak{}TCIA data, including RCC-AID and AIMI annotations. No new participants were recruited or identifiable data collected. Ethical approval was not required for this secondary analysis of public data, as confirmed by the authors.

\section{Acknowledgments and disclosures}
This research received no funding. The authors declare no competing interests.

\end{document}